\documentclass[11pt,a4paper]{article}
\usepackage[hyperref]{emnlp-ijcnlp-2019}
\usepackage{times}
\usepackage{latexsym}
\usepackage{titlesec}
\usepackage{url}
\usepackage{multirow}
\usepackage{graphicx}
\usepackage{amsmath}
\usepackage{CJKutf8}
\usepackage{xcolor}
\usepackage{amssymb}
\usepackage{fixfoot}
\usepackage{caption}
\usepackage{subfigure}
\DeclareFixedFootnote{\BaselineAndTraining}{It is available in the supplemental material.}

\aclfinalcopy 

\title{Improving Bidirectional Decoding with Dynamic Target Semantics in Neural Machine Translation}

\author{
  Yong Shan$^1$$^2$, Yang Feng$^1$$^2$, Jinchao Zhang$^3$, Fandong Meng$^3$, Wen Zhang$^1$$^2$ \\
  $^{1}$ University of Chinese Academy of Sciences \\
  $^{2}$ Key Laboratory of Intelligent Information Processing \\
  Institute of Computing Technology, Chinese Academy of Sciences (ICT/CAS) \\
  $^{3}$ Pattern Recognition Center,	WeChat AI, Tencent Inc, China \\
  {\tt \{shanyong18s,fengyang,zhangwen\}@ict.ac.cn} \\
  {\tt \{dayerzhang,fandongmeng\}@tencent.com}
}

\date{}

\begin{document}
\begin{CJK}{UTF8}{gbsn}
\maketitle
\begin{abstract}
Generally, Neural Machine Translation models generate target words in a left-to-right (L2R) manner and fail to exploit any future (right) semantics information, which usually produces an unbalanced translation. Recent works attempt to utilize the right-to-left (R2L) decoder in bidirectional decoding to alleviate this problem.
In this paper, we propose a novel \textbf{D}ynamic \textbf{I}nteraction \textbf{M}odule (\textbf{DIM}) to dynamically exploit target semantics from R2L translation for enhancing the L2R translation quality.
Different from other bidirectional decoding approaches, DIM firstly extracts helpful target information through addressing and reading operations, then updates target semantics for tracking the interactive history.
Additionally, we further introduce an \textbf{agreement regularization} term into the training objective to narrow the gap between L2R and R2L translations.
Experimental results on NIST Chinese$\Rightarrow$English and WMT'16 English$\Rightarrow$Romanian translation tasks show that our system achieves significant improvements over baseline systems, which also reaches comparable results compared to the state-of-the-art Transformer model with much fewer parameters of it.
\end{abstract}


\section{Introduction}

In recent years, end-to-end neural machine translation (NMT) models  \cite{cho2014learning, sutskever2014sequence,bahdanau2014neural, luong2015effective,wu2016google, gehring2017convolutional, vaswani2017attention} have made promising progress. 
Generally, NMT decoders generate target words in the left-to-right (L2R) manner conditioned on the previously generated words and fail to access the future (right) predictions, which usually leads to the unbalanced translation. 
In other words, the left part of the translation seems more reasonable than the right part. \\
\indent{}To address the aforementioned issue, right-to-left (R2L) decoding are exploited to enhance the L2R decoding procedure.
Rerank-based approaches \cite{liu2016agreement, sennrich2016edinburgh, hoang2017towards} utilize R2L translation models to re-score L2R translations for determining the appropriate translation, but it is difficult for independent decoders to fully exploit target semantics \cite{zhang2018asynchronous}.
Some researchers \cite{yang2018regularizing, zhang2018regularizing, hassan2018achieving} attempt to exploit R2L target semantics through incorporating regularization term into the training objective and neglect deep semantics-level interactions between decoders. 
Recently, \citet{zhang2018asynchronous} propose ABDNMT to employ the attention mechanism for extracting target semantics from hidden states of the R2L decoder to guide the L2R generation.
However, ABDNMT retrieves all R2L hidden states at each step without recording interaction histories, which may lead to inaccurate attention and limited performance. \\
\indent{}In this paper, we propose a novel \textbf{D}ynamic \textbf{I}nteraction \textbf{M}odule (\textbf{DIM}) to extract useful target semantics from the R2L hidden states to guide the L2R generation through addressing and reading operations.
The innovative part in the proposed DIM is the well-designed updating mechanism applied to the R2L hidden states at each decoding step for tracking interaction histories. In this way, the R2L target semantics is dynamically utilized and we expect the DIM to extract more relevant target information with accurate attention distributions. In addition, we integrate an effective \textbf{agreement regularization} based on generated logits of two decoders to narrow the gap between L2R and R2L translations for further enhancing the translation quality. \\
\indent{}We validate the effectiveness of our approach on NIST Chinese$\Rightarrow$English and WMT'16 English$\Rightarrow$Romanian translation tasks. Experimental results show that our approach respectively surpasses the strong NMT baseline system (\textsc{RNNSearch$^*$}) and the strong bidirectional decoding NMT system (\textsc{ABDNMT$^*$}) by \textbf{+3.23} and \textbf{+1.04} BLEU points on the Chinese$\Rightarrow$English task. 
Corresponding improvements on the English$\Rightarrow$Romanian task
are \textbf{+2.25} and \textbf{+1.02} BLEU points, respectively. It is worth mentioning that this is also a comparable result with the state-of-the-art Transformer model but with much fewer parameters of it.
Ablation study further proves the effectiveness of our approach. The attention visualization reflects that the DIM tends to attend the previous un-attended R2L target semantics with more centralized attention distributions. 

\section{Background}

Given a source sentence $\mathbf{x}=\left\{x_1, x_2, \dots, x_n\right\}$ and a target sentence $\mathbf{y}=\left\{y_1, y_2, \dots, y_m\right\}$ , NMT directly models the translation probability as:
\begin{eqnarray}
\begin{split}
p(\mathbf{y}|\mathbf{x};\theta) =& \prod_{t=1}^{m}P(y_t|\mathbf{y}_{<t},\mathbf{x};\theta) \\
=& \prod_{t=1}^{m}\mathrm{softmax}(f(y_{t-1}, \mathbf{c}_t, \mathbf{s}_t))
\end{split} \label{eq_predict}
\end{eqnarray}
where $\theta$ is a set of model parameters, $\mathbf{y}_{<t}=\left\{y_1,\dots,y_{t-1}\right\}$ is a partial translation, $f(\cdot)$ is a non-linear function, and $s_t$ is the hidden state of decoder at time step $t$. $\mathbf{c}_t$ is calculated by the attention mechanism, highly formulated as $\mathbf{c}_t = \mathbf{Attention}(\tilde{\mathbf{s}}_{t-1}, \mathbf{h})$, which is composed of $\mathbf{Address}$ and $\mathbf{Read}$ operations as:
\begin{eqnarray}
\mathbf{a}_t &=& \mathbf{Address}(\tilde{\mathbf{s}}_{t-1}, \mathbf{h}) \\
\mathbf{c}_t &=& \mathbf{Read}(\mathbf{a}_t, \mathbf{h})
\end{eqnarray}
where $\mathbf{a}_t$ is an attention weights calculated by a content-based addressing method, $\mathbf{h}$ is source annotations produced by the encoder and $\tilde{\mathbf{s}}_{t-1}$ is an intermediate state for computing the attention. Refer to \citet{bahdanau2014neural} for details.

\begin{table*}[t!]
\centering
\scalebox{0.80}{
\begin{tabular}{l| l| c | c | cccc | l | c}
\bf \textsc{System}	& \bf \textsc{Architecture} &\bf \# Para. & \bf \textsc{MT02} & \bf \textsc{MT03} & \bf \textsc{MT04} & \bf \textsc{MT05} & \bf \textsc{MT06} & \bf \textsc{Ave.} & \bf $\Delta$ \\
\hline
\citet{bahdanau2014neural} & RNNSearch & 85.62M & -- & 36.59 & 39.57 & 35.56 & 35.29	& 36.75 &  \\
\citet{zhang2018asynchronous} & ABDNMT & 122.86M & 41.60 & 40.02 & 42.32 & 38.84 & 38.38 & 39.89 & \\
\citet{vaswani2017attention} & Transformer & 90.64M & 44.89 & 44.27 & 45.54 & 44.57 & 43.95 & 44.58 &  \\
\hline
\multirow{3}{*}{\em Our Implementations}  			
  &   \textsc{RNNSearch$^{*}$} 	& 46.39M & 44.15 & 42.13  & 43.70  & 41.67  & 42.28  & 42.45 & -- \\
  &   \textsc{ABDNMT$^{*}$}   & 55.84M & 45.77 & 43.91  & 46.15  & 44.07  & 44.44  & 44.64 & +2.19 \\
  &   Our Model 			  & 56.36M & \bf 46.74 & {\bf 45.12}  & {\bf 47.03}  & {\bf 46.20}  & {44.38}  & {\bf 45.68} & \bf +3.23
\end{tabular}
}
\vspace{-0.5em}
\caption{Case-insensitive BLEU scores (\%) on NIST Zh$\Rightarrow$En task. Domain differences between MT02 and MT06 cause our model to decrease slightly over \textsc{ABDNMT$^{*}$} on MT06.} \label{tab_main_zh2en}
\vspace{-6pt}
\end{table*}

\section{Approach}
\subsection{Model Overview}
\begin{figure}[t!]
\begin{center}
\includegraphics[width=0.4\textwidth]{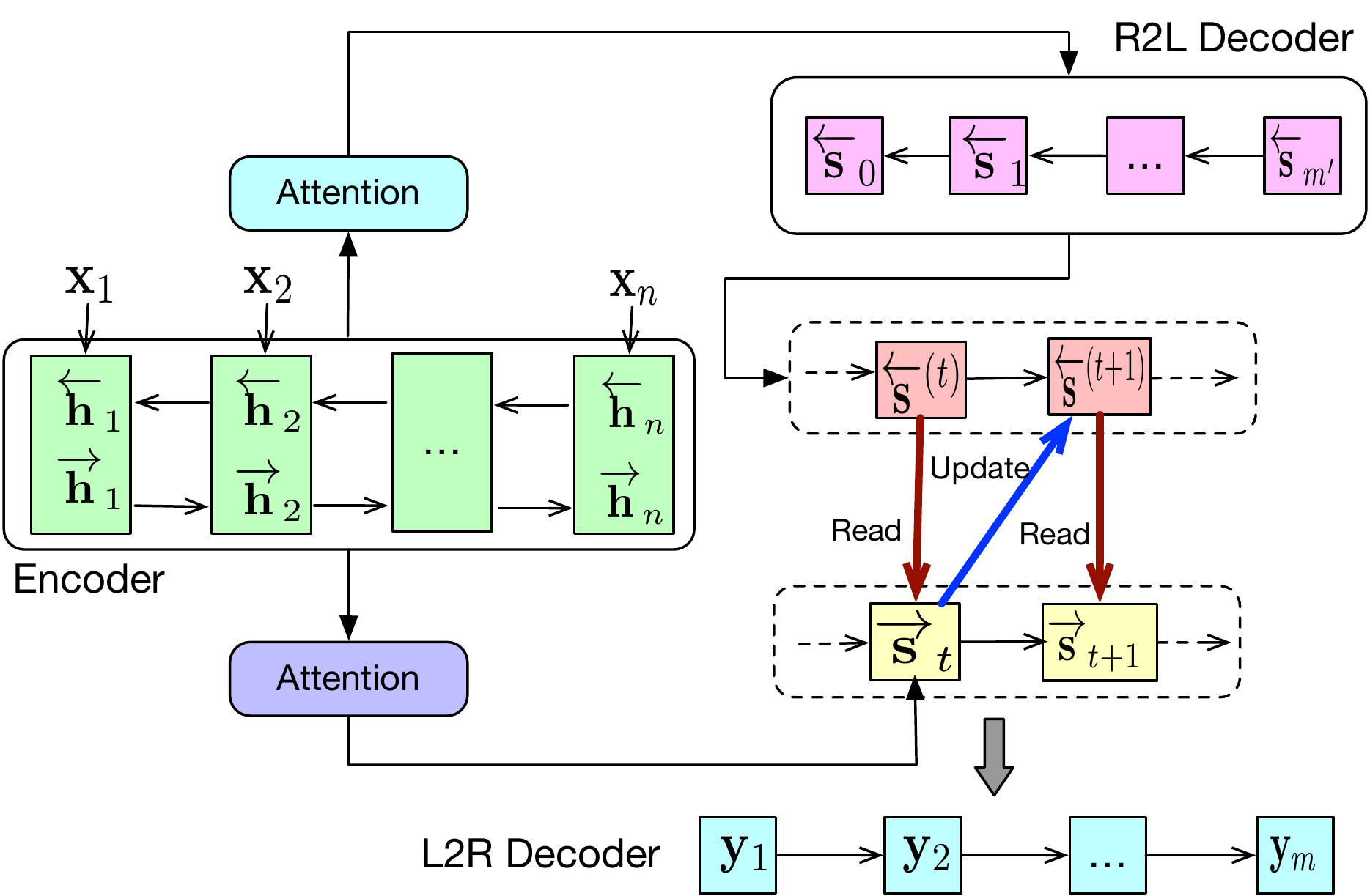}
\vspace{-0.5em}
\caption{The architecture of our proposed model.} \label{img_model} 
\end{center}
\vspace{-8pt}
\end{figure}

As shown in Figure \ref{img_model}, our model mainly includes three components: 1) an encoder with parameter set $\theta^e$ ; 2) a R2L decoder with parameter set $\theta^{r2l}$ ; 3) a L2R decoder with parameter set $\theta^{l2r}$. To be noticed that the L2R decoder generates ultimate translations and the R2L decoder provides target semantics to the L2R decoder.

The R2L decoder generates hidden states $\overleftarrow{\mathbf{s}}=\left\{ \overleftarrow{\mathbf{s}}_1, \overleftarrow{\mathbf{s}}_2, \dots, \overleftarrow{\mathbf{s}}_{m'} \right\}$, where $m'$ indicates the length of translation. It also generates translation $\overleftarrow{\mathbf{y}}=\left\{ \overleftarrow{y}_1, \overleftarrow{y}_2, \dots, \overleftarrow{y}_{m'} \right\}$ to be optimized by a teacher-forcing method at training phase.
Particularly, we use hidden states of the R2L decoder as primary target semantics.
To reduce the time cost, we use greedy search to generate hidden states $\overleftarrow{\mathbf{s}}$.
The L2R decoder generates translation $\overrightarrow{\mathbf{y}}=\left\{\overrightarrow{y}_1, \overrightarrow{y}_2, \dots, \overrightarrow{y}_m \right\}$ where $m$ is the length of translation. 
An attention mechanism is employed to gather related source context $\mathbf{c}_{t}^{l2r}$ according to the L2R decoder's state and source representations at each decoding step $t$, as
$\mathbf{c}_{t}^{l2r}=\mathbf{Attention}(\overrightarrow{\tilde{\mathbf{s}}}_{t-1}, \mathbf{h})$. 
To dynamically exploit target semantics, the L2R decoder employs DIM to extract R2L target semantics from  $\overleftarrow{\mathbf{s}}^{(t)}$ as $\mathbf{c}_t^{tgt}$ and update $\overleftarrow{\mathbf{s}}^{(t)}$ to $\overleftarrow{\mathbf{s}}^{(t+1)}$ at each L2R decoding step, which is described in Section \ref{sub_dim}. $\overleftarrow{\mathbf{s}}^{(t)}$ represents target semantics at decoding step $t$. Then we concatenate the source context and target semantics as $\mathbf{c}_{t}=[\mathbf{c}_{t}^{l2r}; \mathbf{c}_{t}^{tgt}]$ and employ $\mathbf{c}_{t}$ for calculating  $\overrightarrow{\mathbf{s}}_t$ and predicting $\overrightarrow{y}_t$.
\subsection{Dynamic Interaction Module} \label{sub_dim}
Inspired by the neural turing machine \cite{graves2014neural}, the DIM in our model contains three operations: \textbf{Address}, \textbf{Read} and \textbf{Update}. The \textbf{Address} and the \textbf{Read} are the same as they are in the attention mechanism:
\begin{eqnarray}
\mathbf{a}_t^{tgt} &=& \mathbf{Address}(\overrightarrow{\tilde{\mathbf{s}}}_{t-1}, \overleftarrow{\mathbf{s}}^{(t)}) \\
\mathbf{c}_t^{tgt} &=& \mathbf{Read}(\mathbf{a}_t^{tgt}, \overleftarrow{\mathbf{s}}^{(t)}).
\end{eqnarray}
Similar to \citet{meng2016interactive, meng2018neural}, the \textbf{Update} operation consists of two specific actions: \textsc{forget} and \textsc{add}. The \textsc{forget} action erases the redundant content. Specifically, we use $\mathbf{F}_t$ vector to control values to be forgotten on each dimension of $\overrightarrow{\mathbf{s}}_{t-1}$ and use $\mathbf{a}_{t}^{tgt}$ to specifies extents to be erased over different positions of target semantics. Correspondingly, the \textsc{add} action determines the complementary current content to be written into target semantics, which utilizes $\mathbf{A}_t$ vector and $\mathbf{a}_{t}^{tgt}$ to control the values and extents over different positions to be written respectively. The \textbf{Update} operation is formalized by:
\begin{eqnarray}
\overleftarrow{\mathbf{s}}^{(t)} = \underbrace{\overleftarrow{\mathbf{s}}^{(t-1)} \cdot (1-\mathbf{a}_{t}^{tgt} \cdot \mathbf{F}_t)}_{\textsc{forget}} + \underbrace{\mathbf{a}_{t}^{tgt} \cdot \mathbf{A}_t}_{\textsc{add}} \\
\mathbf{F}_t = \sigma(\mathbf{W}_f \overrightarrow{\mathbf{s}}_{t-1}), \hspace{2em} \mathbf{A}_t = \sigma(\mathbf{W}_a \overrightarrow{\mathbf{s}}_{t-1})
\end{eqnarray}
where $\mathbf{W}_f$, $\mathbf{W}_a$ are parameters of \textsc{forget} and \textsc{add} actions and $\sigma$ means the \textit{sigmoid} function.

In this way, the DIM is able to track the interaction histories and dynamically exploit target semantics to extract more relevant target information with accurate attention distributions.

\subsection{Joint Training with Agreement Regularization}
We propose to further introduce joint training with agreement regularization \cite{liang2006alignment, liang2008agreement, cheng2015agreement} into our model. The motivation is to enhance the translation consistency between the R2L decoder and the L2R decoder. We compute the L2 loss between generated logits rather than hidden states \cite{yang2018regularizing} of two decoders as the agreement term due to logits contain more direct prediction information than hidden states. 

Formally, given parameters $\theta^e$, $\theta^{r2l}$, $\theta^{l2r}$ and training data $D=\left\{(\mathbf{x}, \mathbf{y})\right\}$, the training objective function is as follows:
\begin{eqnarray}
\mathcal{L} = \max \frac{1}{\left| D \right|} \sum\limits_{<\mathbf{x}, \mathbf{y}>}
\Big\{
\log(p(\overleftarrow{\mathbf{y}}|\mathbf{x}; \theta^e, \theta^{r2l})) \nonumber\\
+ \log(p(\overrightarrow{\mathbf{y}}|\mathbf{x}; \theta^e, \theta^{r2l}, \theta^{l2r})) - \mathrm{L2}(\overleftarrow{\mathbf{E}}, \overrightarrow{\mathbf{E}})
\Big\} \label{for_loss}
\end{eqnarray}
where $\overleftarrow{\mathbf{E}}$ and $\overrightarrow{\mathbf{E}}$ are generated logits of the R2L decoder and the L2R decoder, respectively. 

\section{Experiments}
\subsection{Setup}
We conduct experiments on NIST Chinese$\Rightarrow$English (Zh$\Rightarrow$En) and WMT'16 English$\Rightarrow$Romanian (En$\Rightarrow$Ro) translation tasks. For Zh$\Rightarrow$En, the training data consists of 1.25M sentence pairs extracted from the LDC corpora. We choose MT02 dataset as the valid set and MT03-MT06 datasets as test sets. For En$\Rightarrow$Ro, we use the preprocessed version of WMT'16 dataset released by \citet{lee2018deterministic} that comprises 0.6M sentence pairs. We use newsdev2016 as the valid set and newstest2016 as the test set. We adopt the \textit{multi-bleu.pl} script to calculate BLEU scores \cite{papineni2002bleu}.

There are three baselines in our experiments: 1) \textbf{\textsc{RNNSearch$^{*}$}}: Our in-house implementation of RNNSearch \cite{bahdanau2014neural}, which is augmented by combining recent advanced techniques\BaselineAndTraining; 2) \textbf{\textsc{ABDNMT$^{*}$}}: Our in-house implementation of ABDNMT \cite{zhang2018asynchronous} based on the strong NMT baseline \textsc{RNNSearch$^{*}$}; 3)\textbf{Transformer}: The state-of-the-art NMT model based on self attention \cite{vaswani2017attention}.
To ensure a fair comparison, we employ the same settings\BaselineAndTraining\ for \textsc{RNNSearch$^{*}$}, \textsc{ABDNMT$^{*}$} and our model. We use base model implemented in fairseq \cite{ott2019fairseq} for Transformer.

\subsection{Results on NIST Chinese$\Rightarrow$English}
Table \ref{tab_main_zh2en} shows results on the NIST Zh$\Rightarrow$En translation task. 
Our model achieves +3.23 BLEU improvements over \textsc{RNNSearch$^*$}, which proves exploiting R2L target semantics can effectively enhance the L2R translation quality.
\textsc{ABDNMT$^*$} is a strong bidirectional decoding baseline that surpasses ABDNMT by +4.75 BLEU on average due to advanced techniques. Our model can still surpass \textsc{ABDNMT$^{*}$} by +1.04 BLEU on average, which proves the effectiveness of our method. Besides, our model surpasses the Transformer model by +1.10 BLEU points with only 63\% parameters compared to it (56.36M vs. 90.64M).

\subsection{Results on WMT'16 English$\Rightarrow$Romanian}
Table \ref{tab_main_en2ro} shows results on WMT'16 En$\Rightarrow$Ro translation task. Our model also outperforms \textsc{RNNSearch$^{*}$} and \textsc{ABDNMT$^{*}$} notably by +2.25 BLEU and +1.02 BLEU respectively. Compared to Transformer, it yields a gain of +0.84 BLEU with only 68\% parameters of it (42.32M vs. 62.05M). It proves that our approach can also achieve remarkable gains in other language pairs.

\begin{table}[h]
\centering
\scalebox{0.80}{
\begin{tabular}{l| l| l| c}
\bf \textsc{Architecture} & \bf \# Para. & \bf \textsc{BLEU} & \bf $\Delta$ \\
\hline
Transformer & 62.05M & 32.85 & \\
\hline
\textsc{RNNSearch$^{*}$} & 33.13M & 31.44 & -- \\
\textsc{ABDNMT$^{*}$} & 41.80M & 32.67 & +1.23 \\
Our Model  & 42.32M & \bf 33.69 & \bf +2.25 \\
\end{tabular}
}
\vspace{-0.5em}
\caption{Case-insensitive BLEU scores (\%) on WMT'16 En$\Rightarrow$Ro test set.} \label{tab_main_en2ro}
\vspace{-8pt}
\end{table}

\begin{figure}[b!]\centering
\vspace{-10pt}
\subfigure[\textsc{ABDNMT$^*$}]{
\includegraphics[width=0.23\textwidth]{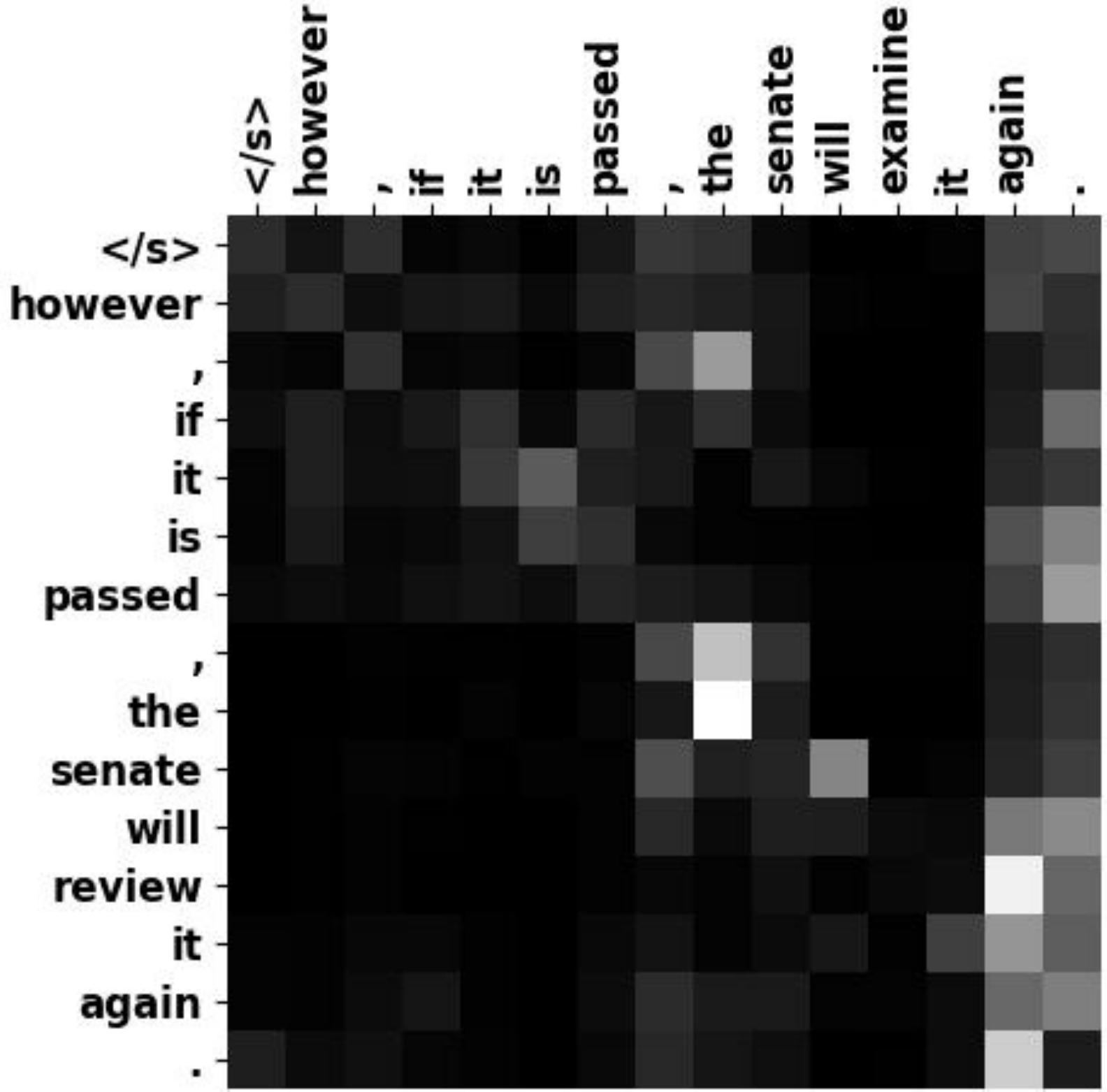}}
\subfigure[Our Model]{
\includegraphics[width=0.23\textwidth]{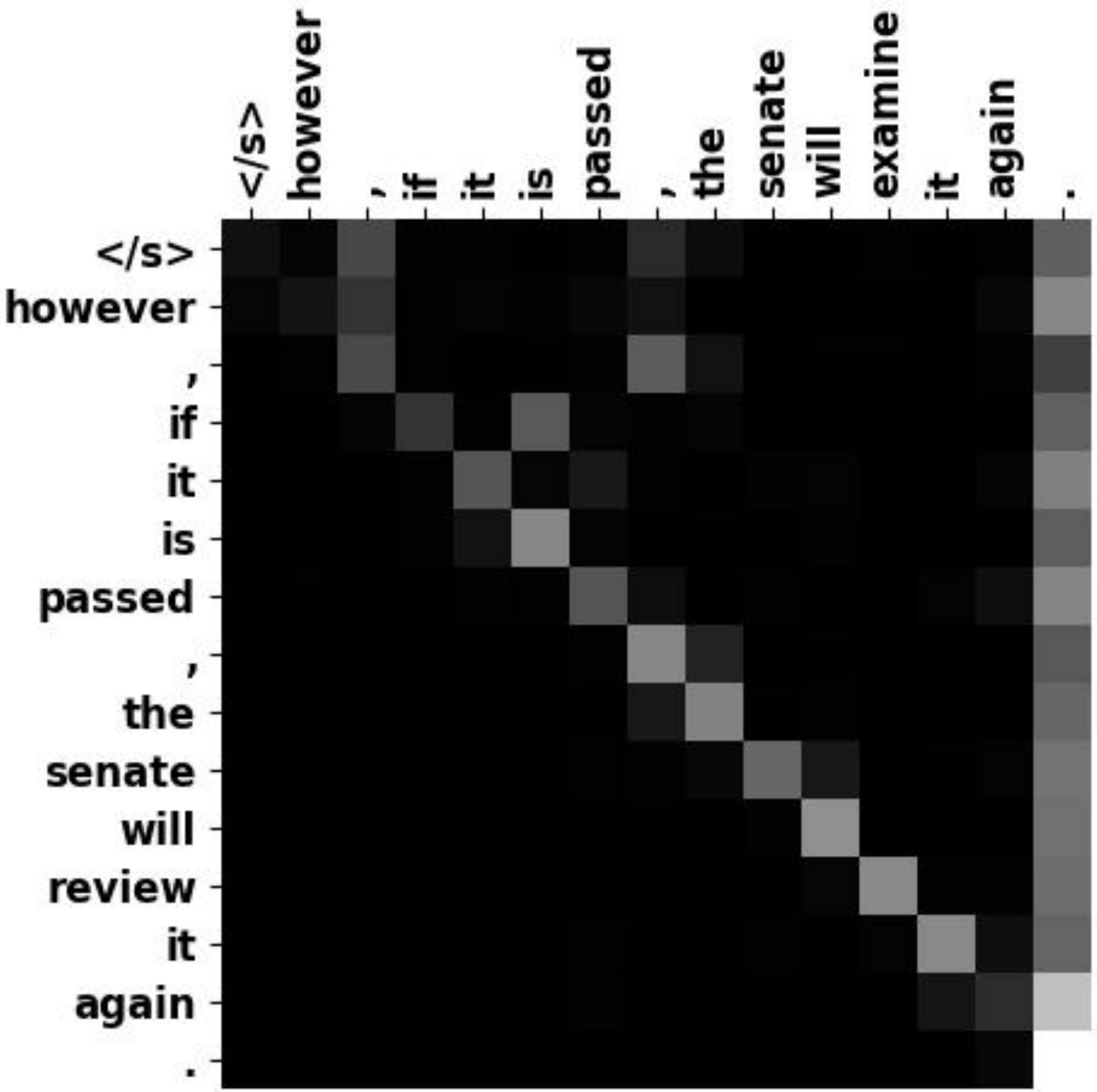}}
\vspace{-1em}
\caption{A visualization example of attention between the R2L decoder and the L2R decoder. The x-axis represents the corresponding R2L translation of target semantics and the y-axis represents the L2R translation. The higher the relevance, the whiter the grayscale.} \label{img_vis_attn}
\end{figure}

\subsection{Ablation Study}
We perform an ablation study on the Zh$\Rightarrow$En valid set to validate the effect of each module. As shown in Table \ref{tab_ablation}, removing agreement regularization or \textbf{Update} operation both decrease the BLEU score but it's still higher than \textsc{ABDNMT$^{*}$}, which proves these two modules are both useful to improve the translation quality. Removing the whole DIM causes a heavy decrease in BLEU scores, which demonstrates the DIM can effectively improve the performance.

\begin{table}[t!]
\scalebox{0.80}{
\begin{tabular}{l | c | c}
\bf \textsc{Architecture} & \bf \textsc{MT02} & \bf $\Delta$ \\
\hline
Our Model & \bf 46.74 & -- \\
~~~~$-$ agreement regularization & 46.39 & -0.35 \\
~~~~$-$ \textbf{Update} & 46.13 & -0.61 \\
~~~~$-$ DIM (\textbf{Address} \& \textbf{Read} \& \textbf{Update}) & 45.15 & -1.59
\end{tabular}
}
\vspace{-0.5em}
\caption{Ablation study of agreement regularization and the DIM on NIST Zh$\Rightarrow$En valid set.} \label{tab_ablation}
\end{table}

\subsection{Attention Visualization}
We choose an example to visualize the attention alignment scores between the R2L decoder and the L2R decoder. As Figure \ref{img_vis_attn} shows, \textsc{ABDNMT$^*$} tends to generate a disperse attention distribution and sometimes attends to the parts that have been translated. However, our model can always produce a high attention score near the diagonal and focus more on the part to translate. We also observe that both models keep focus on the first token of R2L translation ('.' in this example) during the whole decoding procedure, which indicates it has more information to attend.

\subsection{About Length}
\begin{figure}[t!]
\begin{center}
\includegraphics[width=0.4\textwidth]{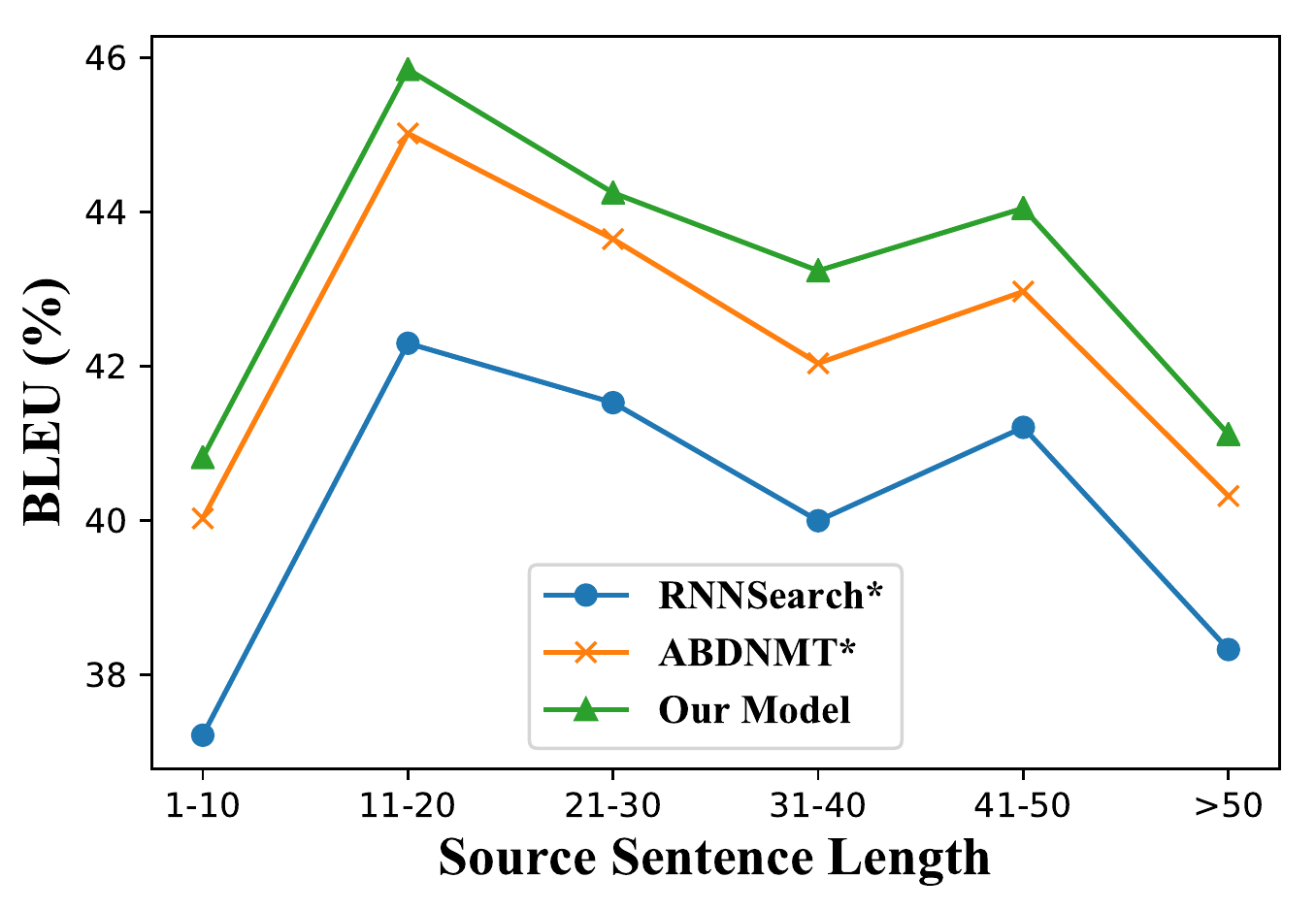}
\vspace{-0.5em}
\caption{The BLEU scores (\%) on the merged test sets with respect to different source sentences lengths.}
\label{img_bleu_per_len} 
\end{center}
\vspace{-10pt}
\end{figure}

To evaluate our model's ability on different sentence lengths, we merge four test sets and test BLEU scores according to different source sentence lengths. As Figure \ref{img_bleu_per_len} shows, our model always yields higher BLEU scores than \textsc{RNNSearch$^{*}$} and \textsc{ABDNMT$^{*}$}, even on long source sentences. It indicates that our method can effectively explore R2L target semantics to improve L2R translation.
\section{Conclusion}
We propose a novel mechanism to improve bidirectional decoding with dynamic target semantics. Experimental results show our method achieves remarkable improvements over our baselines. It also indicates our model can track the interaction histories and lead to more accurate attention distribution. Recently a synchronous decoding approach proposed by \citet{zhang2019synchronous} achieves promising results and we plan to integrate their method into our model for further exploration in the future study.


\bibliography{emnlp-ijcnlp-2019}

\begin{thebibliography}{29}
\expandafter\ifx\csname natexlab\endcsname\relax\def\natexlab#1{#1}\fi

\bibitem[{Bahdanau et~al.(2014)Bahdanau, Cho, and Bengio}]{bahdanau2014neural}
Dzmitry Bahdanau, Kyunghyun Cho, and Yoshua Bengio. 2014.
\newblock Neural machine translation by jointly learning to align and
  translate.
\newblock \emph{arXiv preprint arXiv:1409.0473}.

\bibitem[{Chen et~al.(2018)Chen, Firat, Bapna, Johnson, Macherey, Foster,
  Jones, Parmar, Schuster, Chen et~al.}]{chen2018best}
Mia~Xu Chen, Orhan Firat, Ankur Bapna, Melvin Johnson, Wolfgang Macherey,
  George Foster, Llion Jones, Niki Parmar, Mike Schuster, Zhifeng Chen, et~al.
  2018.
\newblock The best of both worlds: Combining recent advances in neural machine
  translation.
\newblock \emph{arXiv preprint arXiv:1804.09849}.

\bibitem[{Cheng et~al.(2015)Cheng, Shen, He, He, Wu, Sun, and
  Liu}]{cheng2015agreement}
Yong Cheng, Shiqi Shen, Zhongjun He, Wei He, Hua Wu, Maosong Sun, and Yang Liu.
  2015.
\newblock Agreement-based joint training for bidirectional attention-based
  neural machine translation.
\newblock \emph{arXiv preprint arXiv:1512.04650}.

\bibitem[{Cho et~al.(2014)Cho, Van~Merri{\"e}nboer, Gulcehre, Bahdanau,
  Bougares, Schwenk, and Bengio}]{cho2014learning}
Kyunghyun Cho, Bart Van~Merri{\"e}nboer, Caglar Gulcehre, Dzmitry Bahdanau,
  Fethi Bougares, Holger Schwenk, and Yoshua Bengio. 2014.
\newblock Learning phrase representations using rnn encoder-decoder for
  statistical machine translation.
\newblock \emph{arXiv preprint arXiv:1406.1078}.

\bibitem[{Gehring et~al.(2017)Gehring, Auli, Grangier, Yarats, and
  Dauphin}]{gehring2017convolutional}
Jonas Gehring, Michael Auli, David Grangier, Denis Yarats, and Yann~N Dauphin.
  2017.
\newblock Convolutional sequence to sequence learning.
\newblock \emph{arXiv preprint arXiv:1705.03122}.

\bibitem[{Graves et~al.(2014)Graves, Wayne, and Danihelka}]{graves2014neural}
Alex Graves, Greg Wayne, and Ivo Danihelka. 2014.
\newblock Neural turing machines.
\newblock \emph{arXiv preprint arXiv:1410.5401}.

\bibitem[{Hassan et~al.(2018)Hassan, Aue, Chen, Chowdhary, Clark, Federmann,
  Huang, Junczys-Dowmunt, Lewis, Li et~al.}]{hassan2018achieving}
Hany Hassan, Anthony Aue, Chang Chen, Vishal Chowdhary, Jonathan Clark,
  Christian Federmann, Xuedong Huang, Marcin Junczys-Dowmunt, William Lewis,
  Mu~Li, et~al. 2018.
\newblock Achieving human parity on automatic chinese to english news
  translation.
\newblock \emph{arXiv preprint arXiv:1803.05567}.

\bibitem[{Hoang et~al.(2017)Hoang, Haffari, and Cohn}]{hoang2017towards}
Cong Duy~Vu Hoang, Gholamreza Haffari, and Trevor Cohn. 2017.
\newblock Towards decoding as continuous optimization in neural machine
  translation.
\newblock \emph{arXiv preprint arXiv:1701.02854}.

\bibitem[{Kingma and Ba(2014)}]{kingma2014adam}
Diederik~P Kingma and Jimmy Ba. 2014.
\newblock Adam: A method for stochastic optimization.
\newblock \emph{arXiv preprint arXiv:1412.6980}.

\bibitem[{Lee et~al.(2018)Lee, Mansimov, and Cho}]{lee2018deterministic}
Jason Lee, Elman Mansimov, and Kyunghyun Cho. 2018.
\newblock Deterministic non-autoregressive neural sequence modeling by
  iterative refinement.
\newblock \emph{arXiv preprint arXiv:1802.06901}.

\bibitem[{Liang et~al.(2006)Liang, Taskar, and Klein}]{liang2006alignment}
Percy Liang, Ben Taskar, and Dan Klein. 2006.
\newblock Alignment by agreement.
\newblock In \emph{Proceedings of the main conference on Human Language
  Technology Conference of the North American Chapter of the Association of
  Computational Linguistics}, pages 104--111. Association for Computational
  Linguistics.

\bibitem[{Liang et~al.(2008)Liang, Klein, and Jordan}]{liang2008agreement}
Percy~S Liang, Dan Klein, and Michael~I Jordan. 2008.
\newblock Agreement-based learning.
\newblock In \emph{Advances in Neural Information Processing Systems}, pages
  913--920.

\bibitem[{Liu et~al.(2016)Liu, Utiyama, Finch, and Sumita}]{liu2016agreement}
Lemao Liu, Masao Utiyama, Andrew Finch, and Eiichiro Sumita. 2016.
\newblock Agreement on target-bidirectional neural machine translation.
\newblock In \emph{Proceedings of the 2016 Conference of the North American
  Chapter of the Association for Computational Linguistics: Human Language
  Technologies}, pages 411--416.

\bibitem[{Luong et~al.(2015)Luong, Pham, and Manning}]{luong2015effective}
Minh-Thang Luong, Hieu Pham, and Christopher~D Manning. 2015.
\newblock Effective approaches to attention-based neural machine translation.
\newblock \emph{arXiv preprint arXiv:1508.04025}.

\bibitem[{Meng et~al.(2016)Meng, Lu, Li, and Liu}]{meng2016interactive}
Fandong Meng, Zhengdong Lu, Hang Li, and Qun Liu. 2016.
\newblock Interactive attention for neural machine translation.
\newblock \emph{arXiv preprint arXiv:1610.05011}.

\bibitem[{Meng et~al.(2018)Meng, Tu, Cheng, Wu, Zhai, Yang, and
  Wang}]{meng2018neural}
Fandong Meng, Zhaopeng Tu, Yong Cheng, Haiyang Wu, Junjie Zhai, Yuekui Yang,
  and Di~Wang. 2018.
\newblock Neural machine translation with key-value memory-augmented attention.
\newblock \emph{arXiv preprint arXiv:1806.11249}.

\bibitem[{Meng and Zhang(2018)}]{meng2018dtmt}
Fandong Meng and Jinchao Zhang. 2018.
\newblock Dtmt: A novel deep transition architecture for neural machine
  translation.
\newblock \emph{arXiv preprint arXiv:1812.07807}.

\bibitem[{Ott et~al.(2019)Ott, Edunov, Baevski, Fan, Gross, Ng, Grangier, and
  Auli}]{ott2019fairseq}
Myle Ott, Sergey Edunov, Alexei Baevski, Angela Fan, Sam Gross, Nathan Ng,
  David Grangier, and Michael Auli. 2019.
\newblock fairseq: A fast, extensible toolkit for sequence modeling.
\newblock In \emph{Proceedings of NAACL-HLT 2019: Demonstrations}.

\bibitem[{Papineni et~al.(2002)Papineni, Roukos, Ward, and
  Zhu}]{papineni2002bleu}
Kishore Papineni, Salim Roukos, Todd Ward, and Wei-Jing Zhu. 2002.
\newblock Bleu: a method for automatic evaluation of machine translation.
\newblock In \emph{Proceedings of the 40th annual meeting on association for
  computational linguistics}, pages 311--318. Association for Computational
  Linguistics.

\bibitem[{Press and Wolf(2016)}]{press2016using}
Ofir Press and Lior Wolf. 2016.
\newblock Using the output embedding to improve language models.
\newblock \emph{arXiv preprint arXiv:1608.05859}.

\bibitem[{Sennrich et~al.(2015)Sennrich, Haddow, and
  Birch}]{sennrich2015neural}
Rico Sennrich, Barry Haddow, and Alexandra Birch. 2015.
\newblock Neural machine translation of rare words with subword units.
\newblock \emph{arXiv preprint arXiv:1508.07909}.

\bibitem[{Sennrich et~al.(2016)Sennrich, Haddow, and
  Birch}]{sennrich2016edinburgh}
Rico Sennrich, Barry Haddow, and Alexandra Birch. 2016.
\newblock Edinburgh neural machine translation systems for wmt 16.
\newblock \emph{arXiv preprint arXiv:1606.02891}.

\bibitem[{Sutskever et~al.(2014)Sutskever, Vinyals, and
  Le}]{sutskever2014sequence}
Ilya Sutskever, Oriol Vinyals, and Quoc~V Le. 2014.
\newblock Sequence to sequence learning with neural networks.
\newblock In \emph{Advances in neural information processing systems}, pages
  3104--3112.

\bibitem[{Vaswani et~al.(2017)Vaswani, Shazeer, Parmar, Uszkoreit, Jones,
  Gomez, Kaiser, and Polosukhin}]{vaswani2017attention}
Ashish Vaswani, Noam Shazeer, Niki Parmar, Jakob Uszkoreit, Llion Jones,
  Aidan~N Gomez, {\L}ukasz Kaiser, and Illia Polosukhin. 2017.
\newblock Attention is all you need.
\newblock In \emph{Advances in Neural Information Processing Systems}, pages
  5998--6008.

\bibitem[{Wu et~al.(2016)Wu, Schuster, Chen, Le, Norouzi, Macherey, Krikun,
  Cao, Gao, Macherey et~al.}]{wu2016google}
Yonghui Wu, Mike Schuster, Zhifeng Chen, Quoc~V Le, Mohammad Norouzi, Wolfgang
  Macherey, Maxim Krikun, Yuan Cao, Qin Gao, Klaus Macherey, et~al. 2016.
\newblock Google's neural machine translation system: Bridging the gap between
  human and machine translation.
\newblock \emph{arXiv preprint arXiv:1609.08144}.

\bibitem[{Yang et~al.(2018)Yang, Chen, and Le~Nguyen}]{yang2018regularizing}
Zhen Yang, Laifu Chen, and Minh Le~Nguyen. 2018.
\newblock Regularizing forward and backward decoding to improve neural machine
  translation.
\newblock In \emph{2018 10th International Conference on Knowledge and Systems
  Engineering (KSE)}, pages 73--78. IEEE.

\bibitem[{Zhang et~al.(2019)Zhang, Zhou, Zhao, and Zong}]{zhang2019synchronous}
Jiajun Zhang, Long Zhou, Yang Zhao, and Chengqing Zong. 2019.
\newblock Synchronous bidirectional inference for neural sequence generation.
\newblock \emph{arXiv preprint arXiv:1902.08955}.

\bibitem[{Zhang et~al.(2018{\natexlab{a}})Zhang, Su, Qin, Liu, Ji, and
  Wang}]{zhang2018asynchronous}
Xiangwen Zhang, Jinsong Su, Yue Qin, Yang Liu, Rongrong Ji, and Hongji Wang.
  2018{\natexlab{a}}.
\newblock Asynchronous bidirectional decoding for neural machine translation.
\newblock \emph{arXiv preprint arXiv:1801.05122}.

\bibitem[{Zhang et~al.(2018{\natexlab{b}})Zhang, Wu, Liu, Li, Zhou, and
  Chen}]{zhang2018regularizing}
Zhirui Zhang, Shuangzhi Wu, Shujie Liu, Mu~Li, Ming Zhou, and Enhong Chen.
  2018{\natexlab{b}}.
\newblock Regularizing neural machine translation by target-bidirectional
  agreement.
\newblock \emph{arXiv preprint arXiv:1808.04064}.

\end{thebibliography}
\bibliographystyle{acl_natbib}

\appendix
\section{Advanced Techniques}
The encoder of \textsc{RNNSearch$^{*}$} is composed of 2 bidirectional GRUs, totally 4 layer RNNs, and the decoder is composed by conditional GRU, totally 2 layer RNNs. We combine these advanced techniques to enhance our baseline system:
\begin{itemize}
\item \textbf{Multi-head Additive Attention}: A multi-head version of RNNSearch's additive attention.
\item \textbf{Dropout}: We apply dropout on embedding layers, RNN output layers of the encoder, output layer before prediction of the decoder.
\item \textbf{Label Smoothing}: We use uniform label smoothing with an uncertainty=0.1.
\end{itemize}

\section{Training Details}
For Zh$\Rightarrow$En, the number of byte pair encoding (BPE) \cite{sennrich2015neural} merge operations is set to 30k for both source and target languages. For En$\Rightarrow$Ro, we use a shared vocabulary generated by 40k BPEs. The parameters are initialized uniformly between [-0.08, 0.08] and updated by the Adam optimizer \cite{kingma2014adam} ($\beta_1=0.9$, $\beta_2=0.999$ and $\epsilon=1e^{-6}$). And we follow \citet{chen2018best} to vary the learning rate as follows:
\begin{eqnarray}
lr=lr_0 \cdot \mathrm{min} (1+ \frac {t \cdot (n-1)}{np}, n, n \cdot (2n)^{\frac{s-nt}{e-s}}) \nonumber
\end{eqnarray}
where $t$ is the current step, $n$ is the number of concurrent model replicas in training, $p$ is the number of warmup steps, $s$ is the start step of the exponential decay, and $e$ is the end step of the decay. We use 2 Nvidia Titan XP GPUs for synchronous training and set $lr_0$, $p$, $s$ and $e$ to $10^{-3}$, 500, 8000 and 64000 respectively according to \cite{meng2018dtmt}. We train our models on 2 Nvidia Titan XP GPUs and limit tokens per batch to 4096 per GPU. We set \textit{heads count} in multi-head additive attention as 8, word embedding size and hidden size as 512, gradient norm as 5.0. Additionally, we set dropout rates of embedding layers, output layer before prediction of the decoder and RNN output layers of the encoder to 0.5, 0.5 and 0.3 respectively. To reduce GPU memory usage, we tie the word embeddings and output embeddings \cite{press2016using}. During translation phase, we set \textit{beam size} to 10. We use \textit{transformer\_base\_v2} \footnote{https://github.com/tensorflow/tensor2tensor/blob\\/master/tensor2tensor/models/transformer.py} settings to train Transformer in our experiments.

\end{CJK}
\end{document}